\documentclass[conference]{IEEEtran}
\IEEEoverridecommandlockouts
\usepackage{cite}
\usepackage{amsmath,amssymb,amsfonts}
\usepackage{algorithmic}
\usepackage{graphicx}
\usepackage{textcomp}
\usepackage[dvipsnames]{xcolor}
\usepackage{colortbl}
\usepackage{multirow}
\usepackage{booktabs}
\usepackage{enumitem} % 加载 enumitem 宏包
\usepackage{calc}
\usepackage{tikz}
\usepackage{tikz}
\usetikzlibrary{shapes,arrows,positioning,shadows,calc}
\usepackage{xcolor}
\usepackage{mdframed}
\usepackage{threeparttable}
\usepackage{subcaption}  % for subfigure layout
\usepackage{url}

\definecolor{processblue}{RGB}{40,120,181}
\definecolor{connectgreen}{RGB}{35,155,86}
\definecolor{analysispurple}{RGB}{132,89,155}

\def\BibTeX{{\rm B\kern-.05em{\sc i\kern-.025em b}\kern-.08em
    T\kern-.1667em\lower.7ex\hbox{E}\kern-.125emX}}

\begin{document}

\title{Bias in the Shadows: Explore Shortcuts in Encrypted Network Traffic Classification\\
% \title{BiasSeeker: Explore Shortcuts in Encrypted Network Traffic Classification\\
% *\\
}

\author{
\IEEEauthorblockN{Chuyi Wang, Xiaohui Xie\textsuperscript{*} \thanks{\textsuperscript{*} Corresponding Author: Xiaohui Xie (xiexiaohui@tsinghua.edu.cn) and Yong Cui (cuiyong@tsinghua.edu.cn)}, Tongze Wang, Yong Cui\textsuperscript{*}}
\IEEEauthorblockA{\textit{Department of Computer Science and Technology, Tsinghua University}}
}

\maketitle

\begin{abstract}

Pre-trained models operating directly on raw bytes have achieved promising performance in encrypted network traffic classification (NTC), but often suffer from \emph{shortcut learning}—relying on spurious correlations that fail to generalize to real-world data. Existing solutions heavily rely on model-specific interpretation techniques, which lack adaptability and generality across different model architectures and deployment scenarios.

In this paper, we propose BiasSeeker, the first semi-automated framework that is both \emph{model-agnostic} and \emph{data-driven} for detecting dataset-specific shortcut features in encrypted traffic. By performing statistical correlation analysis directly on raw binary traffic, BiasSeeker identifies spurious or environment-entangled features that may compromise generalization, independent of any classifier. To address the diverse nature of shortcut features, we introduce a systematic categorization and apply category-specific validation strategies that reduce bias while preserving meaningful information.

We evaluate BiasSeeker on 19 public datasets across three NTC tasks. By emphasizing context-aware feature selection and dataset-specific diagnosis, BiasSeeker offers a novel perspective for understanding and addressing shortcut learning in encrypted network traffic classification, raising awareness that feature selection should be an intentional and scenario-sensitive step prior to model training.
\end{abstract}


\section{Introduction}
Encrypted Network Traffic Classification~(NTC), which aims to identify categories of network communication patterns, has become an increasingly important research area. This task is critical for ensuring cybersecurity, enhancing service quality and user experience, and enabling efficient network management. 

With the rapid advancement of deep learning (DL), NTC has benefited from a wide range of intelligent methods, including traditional machine learning approaches based on flow statistics~\cite{zhang2014robust, mirsky2018kitsune}, deep learning methods based on flow sequences~\cite{liu2019fs, piet2023ggfast, xue2024fingerprinting}, and pre-trained models operating directly on raw datagram bytes~\cite{lin2022bert, zhao2023yet, wang2024netmamba, peng2024ptu, zhou2024trafficformer}. Among these, pre-trained models have achieved state-of-the-art classification performance. However, they are notably susceptible to the \emph{shortcut learning} problem~\cite{jacobs2022ai, oh2023appsniffer, guthula2023netfound, wickramasinghe2025sok}, wherein models learn to exploit spurious correlations that appear predictive in both training and testing datasets but fail to generalize to real-world, out-of-distribution scenarios~\cite{geirhos2020shortcut}. For instance, pre-trained Transformer models have been observed to rely on TCP Timestamp Options to classify mobile application traffic~\cite{oh2023appsniffer, guthula2023netfound}. 

Existing efforts to prevent shortcut learning in NTC generally fall into two categories:  \emph{model-agnostic interventions} and \emph{model-dependent diagnoses}.  Model-agnostic approaches aim to eliminate widely recognized shortcut features~\cite{guthula2023netfound, wickramasinghe2025sok}, but they often depend heavily on expert domain knowledge and overlook dataset-specific characteristics.
In contrast, model-dependent diagnosis techniques seek to interpret a classifier’s decision-making process to identify shortcuts~\cite{jacobs2022ai, han2024rules}, but their effectiveness is constrained by the particular models and interpretability tools employed. Outside of NTC, the broader AI community has also investigated methods for detecting and mitigating shortcut learning~\cite{ye2024spurious, steinmann2024navigating}, though such techniques are difficult to transfer due to fundamental differences in data modality between binary network traffic and image or text data.

Addressing shortcut learning in NTC faces two core challenges: 
(1)\ \textbf{High diversity in traffic data distributions}. Although prior work attempts to define generic shortcut feature categories, it is extremely difficult to capture the full range of traffic patterns found in diverse environments. For example, traffic collected from backbone networks differs significantly from that of end-user devices in terms of speed, volume, and structure.
(2)\ \textbf{Varying impact of shortcut features across applications}. Previous methods often indiscriminately remove all suspected shortcut features, regardless of context. We argue that shortcut features should instead be carefully categorized and processed based on their intrinsic characteristics to more effectively support robust and generalizable NTC.

% \textcolor{red}{TODO: modify latter paragraphs based on experiment results...}
% In this paper, we introduce \textbf{BiasSeeker}, the model-agnostic and data-driven framework for detecting dataset-specific shortcut features in encrypted network traffic classification. 
% To accurately capture the intrinsic distribution of each dataset, BiasSeeker operates directly on raw binary traffic and employs statistical correlation analysis to systematically identify spurious features that are strongly associated with class labels. Its model-agnostic nature enables BiasSeeker to detect a wide range of shortcut features without being limited by any particular classifier or interpretation technique.
% To better mitigate the effects of dataset-specific shortcuts, we further propose a principled categorization of known shortcut features and develop corresponding category-specific mitigation strategies. This ensures that useful traffic information is preserved while harmful shortcuts are suppressed, thereby encouraging models to learn more stable and semantically meaningful representations.
% We validate the effectiveness of BiasSeeker through extensive experiments on 19 public traffic datasets spanning three major classification tasks, demonstrating its strong performance in both shortcut detection and mitigation. As the first data-driven framework tailored specifically to shortcut learning in NTC, BiasSeeker offers new insights into the underlying mechanisms of learning-based traffic classification.

In this paper, we introduce \textbf{BiasSeeker}, the first semi-automated framework that is both model-agnostic and data-centric, offering a new perspective on shortcut learning in encrypted network traffic classification.  Operating directly on raw binary traffic, it employs statistical analysis to detect spurious or environment-entangled features that may compromise generalization.

To promote meaningful learning and mitigate shortcut reliance, we further propose a principled taxonomy of shortcut feature types and develop corresponding validation strategies tailored to their behavioral patterns. These approaches aim not to eliminate features indiscriminately, but to retain valuable traffic information while suppressing dataset-induced biases. Feature selection should thus be an intentional and scenario-sensitive step prior to model training or benchmarking. 

Our evaluations span 19 publicly available datasets across VPN, malware, and encrypted application classification tasks, providing empirical evidence of BiasSeeker's effectiveness in detecting shortcut features. 
By emphasizing context-aware feature selection and dataset-specific diagnosis, BiasSeeker contributes a data-centric foundation for building more robust, generalizable, and trustworthy network traffic classifiers.

Our main contributions are summarized as follows:
\begin{itemize}
    \item We introduce a model-agnostic, data-specific perspective that detects feature instabilities through a mathematically-driven semi-automated framework combining statistical analysis.
    \item We develop a systematic categorization of feature instability patterns and design targeted mitigation strategies that preserve valuable traffic information while addressing specific types of unreliable feature dependencies.
    \item We conduct comprehensive evaluations across 19 diverse datasets spanning VPN, malware, and encrypted application classification tasks, empirically validating the effectiveness of our approach in detecting and mitigating unstable feature behaviors.
    \item We propose actionable insights and research directions about the development of accurate, generalizable, and real-world applicable network traffic classifiers through data-centric resolution.
\end{itemize}

% \textcolor{red}{TODO: modify latter paragraphs based on chapter structure}
% The rest of the paper is organized as follows. Section~\ref{sec:related} reviews related work on feature instability and encrypted traffic classification. Section~\ref{sec:prelim} introduces the preliminary definitions and problem setup. 
% Section~\ref{sec:method} details our proposed BiasSeeker framework, develops the categorization and devises category-specific validation strategies.
% Section~\ref{sec:experiments} presents the experimental settings. Section~\ref{sec:result} discusses our empirical results and analyses. Finally, Section~\ref{sec:conclusion} concludes the paper and outlines future research directions.

\section{Related Works}\label{sec:related}

\subsection{Shortcut Learning in AI Community}
In classification tasks, the tendency of models to rely on spurious correlations between input data and ground truth labels, rather than meaningful features, are referred to as shortcuts~\cite{geirhos2020shortcut}.
To prevent models from relying on such shorts, researchers have explored a variety of model-dependent detection methods. 

Assuming that shortcuts are easier to learn than the relevant features, Nam et al.~\cite{nam2020learning} trained an auxiliary detector to identify shortcut samples by increasing their loss gradients. Similarly, Spare~\cite{yang2024identifying} detects shortcut samples by applying clustering techniques during the early stages of training. 

Inspired by perturbation-based feature importance, Wang et al.~\cite{wang2023neural} assesses the importance of frequency-based feature by sequentially removing specific frequency bands, revealing shortcuts in image classification through changes in model performance.

To identify spurious features based on domain knowledge, researchers have also adopted techniques from eXplainable Artificial Intelligent~(XAI). In addition to widely used libraries such as InterpretML~\cite{nori2019interpretml} and Captum~\cite{kokhlikyan2020captum}, more advanced methods such as heatmap clustering~\cite{lapuschkin2019unmasking} or feature disentanglement~\cite{muller2024shortcut} have also been proposed.

Departing from the focus on textual or visual modalities, our work explores the detection of potential shortcut features specifically within network traffic data.

\subsection{Shortcut Learning in NTC}
To develop more robust network traffic classification models, researchers have explored two main approaches to mitigate shortcut learning: \emph{model-agnostic prior intervention} and \emph{model-dependent post-hoc diagnosis}.  

Model-agnostic prior intervention methods aim to remove features that are commonly assumed to be shortcuts. For example, YaTC~\cite{zhao2023yet} and NetMamba~\cite{wang2024netmamba} remove Ethernet header and IP addresses; ET-BERT~\cite{lin2022bert} excludes both Ethernet and IP headers; and netFound~\cite{guthula2023netfound} and TrafficFormer~\cite{zhou2024trafficformer} go further by removing or randomizing fields such as sequence numbers, acknowledgment numbers, TCP timestamp options, and server name indication (SNI) in TLS handshake messages. More recently, NTC-Enigma~\cite{wickramasinghe2025sok} provides a comprehensive summary of these shortcut-related fields based on RFC specifications. 
However, all of these approaches rely on predefined assumptions about which features constitute shortcuts, which may not accurately reflect the actual feature distributions of a given dataset. This limitation motivates a data-driven approach to detecting shortcuts more precisely.

Conversely, model-dependent post-hoc diagnosis methods seek to interpret a classifier’s decision-making process or analyze feature importance to identify and mitigate shortcuts. For example, Trustee~\cite{jacobs2022ai} identifies shortcuts by distilling black-box models into interpretable ones such as random forests. However, the effectiveness of these methods is constrained by the specific classifiers and interpretation techniques used. Furthermore, explanations provided by interpretable models are often not inherently easier to understand~\cite{mink2023everybody}. 
This underscores the need for a model-agnostic method with greater potential to reveal dataset-level biases in a more general and scalable manner.

In contrast to these existing efforts, our work introduces a data-driven and model-agnostic framework for detecting potential shortcut features in network traffic.

\section{Preliminaries} \label{sec:prelim}
In this section, we introduce the theoretical foundations and statistical criteria underpinning our framework for shortcut detection in encrypted traffic classification. Our goal is to identify features whose apparent correlation with class labels arises from dataset-specific artifacts—rather than genuine protocol semantics—and which thus pose risks to model generalization.

\subsection{Shortcut Definition}
In machine learning, \emph{shortcut learning} refers to the phenomenon where models rely on non-causal signals to make predictions~\cite{geirhos2020shortcut}. These shortcuts typically emerge from incidental patterns in training data that are predictive of labels but do not reflect true task semantics. Models exploiting shortcuts perform well in-distribution but generalize poorly under domain shifts.

In encrypted network traffic classification, shortcut learning manifests uniquely. With packet payloads inaccessible, models rely solely on side-channel information such as headers, timing patterns, or flow metadata. When header fields or timing signals correlate with labels due to capture artifacts or protocol idiosyncrasies—rather than intrinsic traffic characteristics—they become shortcuts. For example, TCP timestamp options may uniquely identify mobile applications in specific datasets but fail when timestamps are randomized or disabled~\cite{oh2023appsniffer}. Similarly, IP addresses may appear predictive in static network environments but become unreliable under NAT or dynamic addressing~\cite{guthula2023netfound}. These shortcuts create brittle models that fail in real-world deployments.

\subsection{Problem Formulation}
Let $\mathcal{D} = \{(x_i, y_i)\}_{i=1}^N$ denote a dataset, where $x_i \in \mathbb{R}^d$ is a $d$-dimensional feature vector and $y_i \in \mathcal{Y}$ its label. Let $x_i^{(j)}$ denote the $j$-th feature of the $i$-th sample, and $X_j = \{x_i^{(j)}\}_{i=1}^N$ the empirical realization of feature $j$. Our goal is to quantify the statistical dependence between $X_j$ and labels $Y = \{y_i\}_{i=1}^N$, and assess whether this dependence likely arises from genuine semantics or potential shortcut mechanisms.

\subsection{Mutual Information (MI)}
For discrete-valued feature $X_j$ and categorical label $Y$, mutual information quantifies the reduction in uncertainty about one variable given knowledge of the other:
\[
I(X_j; Y) = \sum_{x \in \mathcal{X}_j} \sum_{y \in \mathcal{Y}} P(x, y) \log \left( \frac{P(x, y)}{P(x)P(y)} \right),
\]
where $P(x, y)$ is the joint empirical distribution and $P(x), P(y)$ the marginals.

High MI indicates that $X_j$ provides substantial information about $Y$. In traffic classification, this suggests strong predictiveness, potentially from semantic patterns or spurious correlations.

However, MI is sensitive to feature cardinality and label imbalance, leading to overestimation of dependence. It also lacks a natural upper bound, complicating cross-feature comparisons.

\subsection{Adjusted Mutual Information (AMI)}
To address these limitations, we use \emph{Adjusted Mutual Information (AMI)}, which normalizes MI against its expected value under a random null model and bounds the score between 0 and 1.

For discrete variables $X_j$ and $Y$:
\[
\mathrm{AMI}(X_j, Y) = \frac{I(X_j; Y) - \mathbb{E}[I(X_j; Y)]}{\max\{H(X_j), H(Y)\} - \mathbb{E}[I(X_j; Y)]},
\]
where $H(\cdot)$ is Shannon entropy, and $\mathbb{E}[I(X_j; Y)]$ is the expected MI under random permutation. AMI offers key advantages:
\begin{itemize}
    \item Normalization: Bounded in $[0, 1]$, enabling fair comparison across features.
    \item Adjustment for Chance: Discounts spurious agreement from random correlations.
    \item Robustness: Less sensitive to class imbalance, suitable for traffic classification.
\end{itemize}

\section{Methodology} \label{sec:method}

In this section, we present our integrated framework for detecting, categorizing, and further validating shortcut features in encrypted traffic classification. Our approach combines data-driven statistical analysis with domain knowledge to systematically identify problematic features and develop targeted mitigation strategies.

\subsection{Intuition}
Our intuition is that shortcut features tend to exhibit disproportionately high statistical association with the label. Thus, we first identify the most label-correlated features from the full feature set as candidates for further examination.

We identify a feature as a \emph{potential shortcut} if:
\begin{enumerate}
    \item It exhibits a high AMI score with respect to the class label, indicating strong predictive power;
    \item Its semantics suggest no causal or protocol-level connection to the class (e.g., hardware identifiers, checksum fields, timestamp offsets);
    \item It shows fragile behavior across environments or datasets, which we investigate further in Section~\ref{sec:experiments}.
\end{enumerate}

By systematically auditing such features, we aim to reduce shortcut learning and promote generalizable, protocol-aware classification.

\subsection{Overview}
The overall workflow of our framework includes:

\begin{enumerate}
    \item Extraction of all raw packet fields using advanced packet parsing tools.
    \item Preprocessing and conversion of raw fields into a uniform integer-based format.
    \item Calculation of Adjusted Mutual Information (AMI) between each feature and the class label.
    \item Selection of the top-$k$ features ranked by AMI as candidate shortcut features.
    \item Categorization of these features based on domain knowledge into three types.
    \item Category-specific validation and mitigation strategies (detailed in Section~\ref{sec:experiments}).
\end{enumerate}

\begin{figure}[htbp]
    \centerline{\includegraphics[scale=0.27]{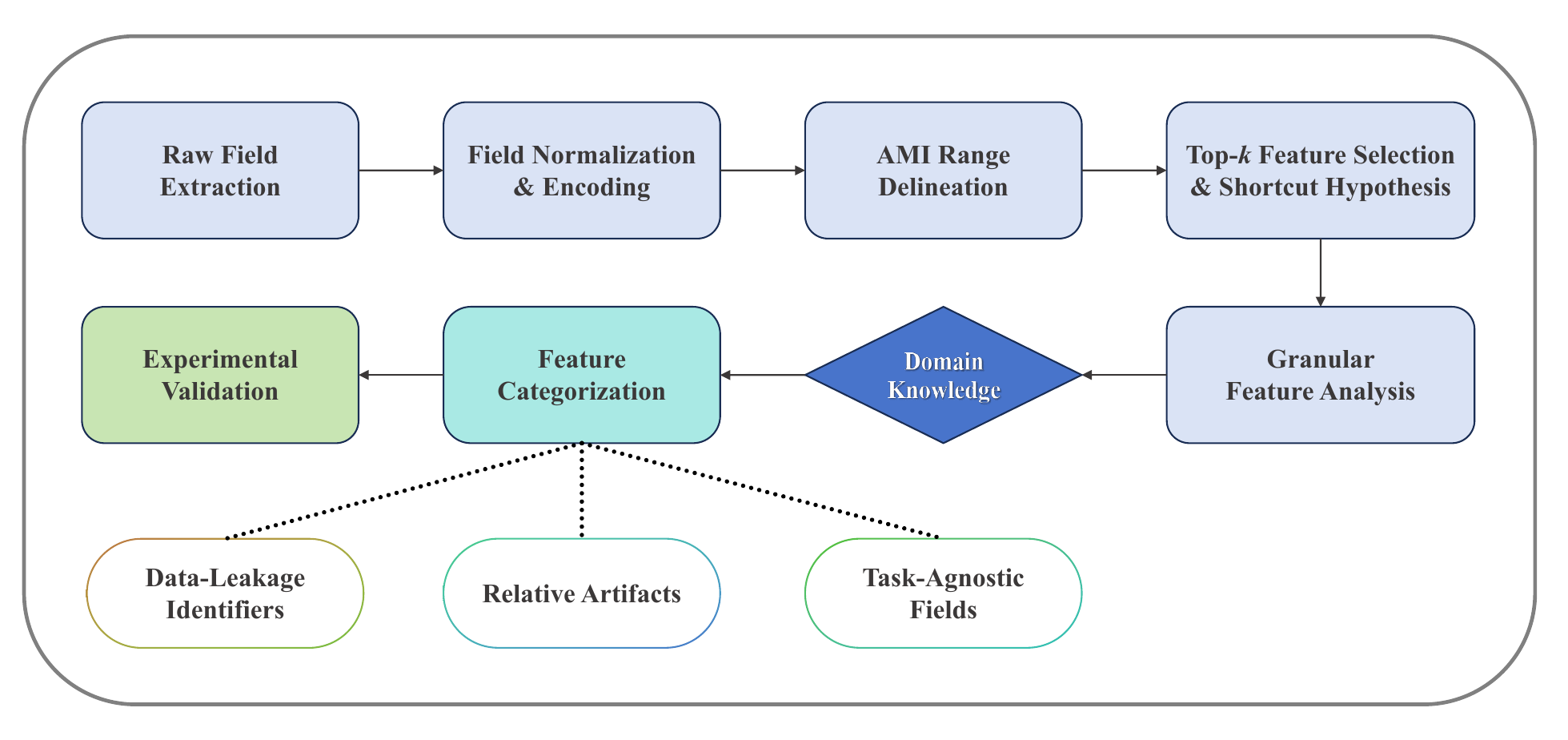}}
    \caption{Workflow of the Feature Analysis Process}
    \label{fig:method}
\end{figure}

\subsection{Pipeline}
The workflow of our framework shown in Fig~\ref{fig:method}is designed to systematically analyze, categorize, and mitigate shortcut features. We now detail each component.

\subsubsection{Raw Field Extraction} 
We utilize \textit{tshark}\footnote{\url{https://tshark.dev/}} to extract full packet-level fields, owing to its extensive protocol support, structured output, and community-maintained dissectors. \texttt{tshark} built atop Wireshark, offers the richest set of fields and the most precise dissection of encrypted traffic layers.

\subsubsection{Field Normalization and Encoding} 
To compute statistical measures such as AMI, all fields must be transformed into numerical formats. Our preprocessing pipeline includes unified packet-level feature extraction and encoding routines, designed to handle a wide range of protocol layers and field types:

\begin{itemize}
    \item IP Address Conversion: IP addresses are converted into 32-bit integers by treating each octet as a base-256 digit.
    \item Hexadecimal Decoding: Fields are parsed from hexadecimal or string representations to integers using robust fallback conversion routines.
    \item Floating-Point Preservation: Temporal fields are retained as floating-point values to preserve fine-grained resolution.
    \item Domain Name Normalization: For TLS/DTLS server names, we extract second-level domains and map them to consistent integer indices using a global dictionary constructed during parsing.
    \item Missing Value Handling: Empty or structurally invalid values are assigned a default excluded value to prevent NaNs or spurious outliers from affecting downstream statistics.
\end{itemize}

To ensure data quality, we filter out redundant or incomplete traffic flows, such as samples with less than 5\% valid fields or malformed IP address mappings.

\subsubsection{AMI Range Delineation} 
We assess the informativeness of each feature using Adjusted Mutual Information (AMI) between the feature and the classification label. 
% Let $X_j$ denote the $j$-th feature and $Y$ the label. 
% AMI is defined as:

% \[
% \text{AMI}(X_j, Y) = \frac{I(X_j; Y) - \mathbb{E}[I(X_j; Y)]}{\max\{H(X_j), H(Y)\} - \mathbb{E}[I(X_j; Y)]}
% \]

% where $I(X_j; Y)$ denotes mutual information and $H(\cdot)$ is the entropy. The expectation $\mathbb{E}[I(X_j; Y)]$ is computed under the null hypothesis that $X_j$ and $Y$ are statistically independent.

% AMI possesses several key properties that make it suitable for detection:

% \begin{itemize}
%     \item \textbf{Boundedness}: $0 \leq \text{AMI}(X_j, Y) \leq 1$, where 0 indicates no association and 1 indicates perfect association.
%     \item \textbf{Invariance}: AMI is invariant to permutations of the label or feature values.
%     \item \textbf{Asymptotic Consistency}: As sample size increases, AMI converges to the population value.
%     \item \textbf{Chance Correction}: AMI accounts for the expected agreement under random assignment, reducing false positives.
% \end{itemize}

To ensure reliable estimates:
\begin{itemize}
    \item Categorical fields are encoded using \texttt{LabelEncoder}.
    \item Numerical fields are discretized if their distributions are highly skewed or multimodal.
\end{itemize}

Prior to AMI computation, we remove constant fields, low-entropy fields, and structurally trivial attributes (e.g., frame number, stream index, Ethernet padding) to avoid misleadingly high or spurious AMI scores.

\subsubsection{Top-$k$ Feature Selection and Shortcut Hypothesis} 
We rank features by their AMI scores and retain the top-ranked for further inspection. We formally justify that high-AMI features are necessary (though not sufficient) conditions for shortcut presence:

If a feature $X_j$ is exploited as a shortcut by a model $f$, then $X_j$ must exhibit non-trivial dependence with the label $Y$, i.e., $I(X_j; Y) > 0$. Therefore, shortcut features are expected to appear in the top-ranked AMI list.

This provides a sound basis for focusing on top-AMI features in the initial detection phase.

\subsection{Categorization}
Network traffic classification (NTC) models often rely on unintended correlations between features and labels, these issues vary across tasks and datasets. We introduce a method to categorize such features based on domain knowledge. Our categorization framework identifies three distinct types of potential shortcut features:

\begin{itemize}
    \item \textbf{Data-Leakage Identifiers}: Features that reveal label-relevant information due to data collection or labeling artifacts.
    
    \item \textbf{Relative Artifacts}: Fields that contain absolute values encoding host- or session-specific behavior, which, though not explicitly leaking labels, produce learnable patterns irrelevant to true semantics.
    
    \item \textbf{Task-Agnostic Fields}: Low-level protocol features that correlate with environmental conditions rather than application behavior. Their shortcut effect emerges from spurious alignment with label distributions across datasets.
\end{itemize}

This categorization enables targeted analysis and validation strategies for each type of potential shortcut feature, as detailed in the following subsection.

\subsection{Validation}
% model-based 验证，三种 zero padding, relative , random mask
% 三个类别 zero padding, relative , random
% evaluation strategies 

% category-specific 方法，KL 散度和 KDE
% relative artifacts
% task-agnostic fields

To support our categorization and better understand the shortcut behaviors of different feature types, we design two complementary validation strategies: model-based validation and category-specific analyses. Together, these approaches verify the practical impact of suspected shortcut features and inform mitigation design.

\subsubsection{Model-Based Validation}

To quantify how shortcut-prone features influence downstream classification performance, we apply three occlusion strategies at the feature level: \emph{zero padding}, \emph{relative transformation}, and \emph{random masking}. These strategies are designed to suppress shortcut signals while preserving feature structure as much as possible.

% \begin{itemize}
%     \item \textbf{Zero Padding:} The simplest method, replacing the target feature with zeros to remove its influence while maintaining input shape.
%     \item \textbf{Relative Transformation:} Designed for features encoding host- or session-specific absolute values. Instead of raw values, we compute their deltas between adjacent packets, preserving temporal semantics while removing static bias.
%     \item \textbf{Random Masking:} Provides a stronger obfuscation mechanism, randomizes values in the first packet, and propagates transformations to maintain consistency within the session.
% \end{itemize}

These strategies are later applied in Section~\ref{sec:experiments} to assess how performance shifts when suspected shortcut features are occluded or transformed, offering an indirect yet practical measurement of their influence.

\subsubsection{Category-Specific Analyses}

Beyond model-agnostic occlusion, we develop tailored validation methods that exploit the intrinsic statistical properties of feature categories.

\paragraph{Relative Artifacts}  
To validate whether absolute-valued features encode spurious host/session information, we measure the change in feature-label dependence before and after relative transformation using Adjusted Mutual Information (AMI):
\[
\Delta_{\text{AMI}} = \text{AMI}(X_j, Y) - \text{AMI}(X_j^{\text{rel}}, Y)
\]
A significant drop in AMI score indicates that absolute values contribute little beyond host-specific bias, supporting the shortcut hypothesis. This analysis justifies using relative encoding as a mitigation strategy.

\paragraph{Task-Agnostic Fields}  
These low-level protocol fields often correlate with environment-specific properties. We quantify their cross-dataset generalizability using class-conditional Kullback-Leibler divergence:
\[
\text{KL}_{\text{avg}}(X_j) = \frac{1}{|C|} \sum_{y \in C} \text{KL}(P_{X_j}^{(D_1)}(y) \Vert P_{X_j}^{(D_2)}(y))
\]
where $P_{X_j}^{(D)}(y)$ is the empirical distribution of feature $X_j$ under class $y$ in dataset $D$. High divergence suggests dataset entanglement and poor transferability. Distribution-aware normalization may reduce this bias.

The category-specific analyses ensure that our categorization is not only theoretically sound but also practically actionable, providing targeted validation strategies for each type of potential shortcut feature. Feature selection is grounded not only in intuition but also in quantifiable evidence.

\section{Experiments}\label{sec:experiments}

\subsection{Experimental Settings}

\begin{figure*}[htpb]
    \centering
    \includegraphics[scale=0.29]{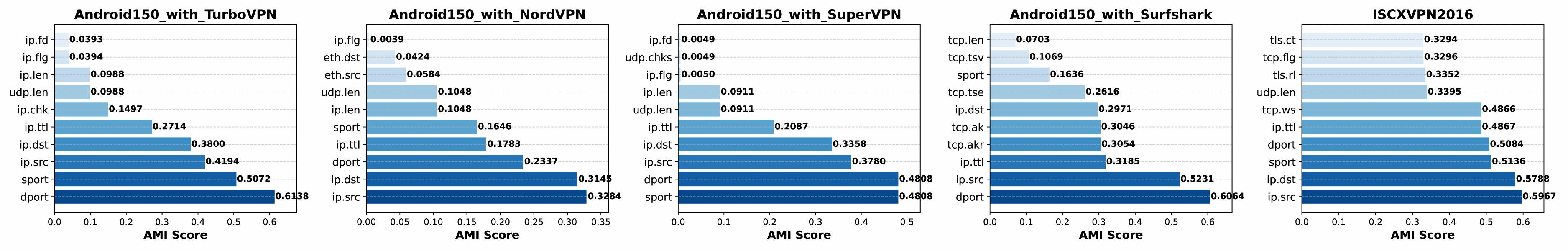}
    \caption{Top-k AMI Features in VPN Traffic Datasets}
    \label{fig:vpn}
\end{figure*}

% \begin{figure*}[htpb]
%     \centering
%     \includegraphics[scale=0.28]{pics/Malware.pdf}
%     \caption{Malware Traffic Datasets}
%     \label{fig:malware}
% \end{figure*}

\subsubsection{Datasets}
To evaluate the effectiveness of the proposed approach, we perform experiments on a diverse set of publicly available traffic datasets, covering three primary encrypted traffic classification tasks.

\begin{enumerate}[label=\arabic*.]
\item \textbf{VPN Traffic Classification:} This task involves identifying different categories of network traffic routed through VPN services. Specifically, datasets such as NordVPN \cite{oh2023appsniffer}, SuperVPN \cite{oh2023appsniffer}, Surfshark \cite{oh2023appsniffer}, TurboVPN \cite{oh2023appsniffer} include traffic captured from 100 Android mobile applications. In addition, ISCXVPN2016 \cite{gil2016characterization} contains both VPN and non-VPN traffic collected in real-world scenarios, covering diverse activities such as web browsing, email communication, and streaming.
\item \textbf{Malware Traffic Classification:} The goal is to distinguish malware-generated traffic from benign traffic. We use all four malware categories from CIC-AndMal2017~\cite{HabibiLashkari2018TowardDA}, namely Adware, Scareware, SMSMalware and Ransomware. Additionally, we employ the USTC-TFC2016 \cite{wang2017malware}, which comprises both benign and malicious traffic samples.
% , which contain encrypted flows of various mobile and desktop applications collected under diverse encryption protocols.
\item \textbf{Encrypted Application Classification:} This task aims to classify encrypted traffic according to the originating applications. The CrossPlatform(Android)\cite{ren2019international} and CrossPlatform(iOS)\cite{ren2019international} contain traffic traces from over 200 mobile applications. CrossNet2021\cite{li2022robust} captures traffic from the same applications under varying network conditions. Both CSTNET-TLS1.3\cite{lin2022bert} and CipherSpectrum \cite{wickramasinghe2025sok} feature application-level traffic encrypted using the TLS 1.3 protocol.
\end{enumerate}

\subsubsection{Shortcut Detection Setting}
To uncover shortcuts in each dataset, we compute AMI in a per-packet granularity. Our choice to use packet-level features is driven by the desire to maintain high fidelity and alignment across diverse datasets. Shortcut signals often reside in individual packet-level metadata, such as specific TCP flags, TTL values, or timestamp behaviors. While flow-level aggregation could capture long-term dependencies, it also introduces challenges: flow lengths vary greatly, making truncation or padding necessary, which often leads to bias or information loss. Flow-level statistics in shortcut detection may obscure such subtle yet discriminative patterns. Therefore, a per-packet approach allows us to retain and highlight these fine-grained correlations. 
% Nonetheless, we are actively investigating hybrid strategies that balance temporal coverage and robustness.

\subsubsection{Model Selection}
To assess the impact of the identified shortcuts on model classification performance, we employ two types of test models: (1) \textbf{NetMamba}~\cite{wang2024netmamba}, a state-of-the-art pre-trained model for network traffic classification, and (2) a \textbf{decision tree}, representing a classical shallow machine learning approach. The input data is prepared following NetMamba's original design: for each flow, we select the first 5 packets, extracting the first 80 header bytes and first 240 payload bytes. The main differences concerning data representation are two-folds:
\begin{enumerate}[label=\arabic*.]
    \item We replace the uni-directional flow with bi-directional session flows.
    \item We keep or discard certain header fields according to our mitigation strategies.
\end{enumerate}

We pre-train NetMamba on 6 public datasets: CICIoT2022\cite{dadkhah2022towards}, CrossPlatform(Android), CrossPlatform(iOS), ISCXVPN2016, USTC-TFC2016 and ISCXTor2016\cite{lashkari2017characterization}.

To reduce resource costs for model fine-tuning and mitigation evaluation, we select two datasets from each classification task. For datasets with class numbers more than 30 (i.e., CSTNET-TLS1.3 and SurfsharkVPN), we randomly sampled a small proportion of original categories. For each dataset, we randomly sample at most 500 flows. All flows in each category are divided into training/validation/test in an 8:1:1 ratio. To mitigate randomness, we repeat the sampling process three times per dataset and report the average model performance across these runs.

Detailed statistics of datasets used for shortcut mitigation are outlined in Table~\ref{tab:mitigation_dataset}.

\begin{table}[htpb]
    \centering
    \scriptsize
    \caption{The Statistics of Shortcut Mitigation Datasets}
    \label{tab:mitigation_dataset}
    \begin{tabular}{c|c|m{0.6cm}m{0.6cm}m{0.6cm}m{0.6cm}}
    \toprule
    Task & Dataset & \#Class & \#Flows & \#Used Class & \#Used Flows \\
    \midrule
\multirow{2}{*}{Application} & CrossNet2021 & 20 & 4443 & 20 & 4443 \\\cmidrule(lr){2-6}
    & CSTNET-TLS1.3 & 120 & 46372 & 18 & 7108 \\\midrule
\multirow{2}{*}{VPN} & ISCXVPN2016 & 13 & 37770 & 13 & 6124 \\\cmidrule(lr){2-6}
    & SurfsharkVPN & 151 & 7500 & 30 & 1500 \\\midrule
\multirow{2}{*}{Malware} & USTC-TFC2016 & 20 & 489139 & 20 & 10000 \\\cmidrule(lr){2-6}
    & Ransomware & 10 & 196803 & 10 & 5000 \\
    \bottomrule
    \end{tabular}
\end{table}

\subsubsection{Model-based Validation Strategies}
To assess the influence of suspicious shortcut features on a given NTC model, we design three feature occlusion strategies: \emph{zero padding}, \emph{relative transformation}, and \emph{random masking}.

Zero padding is the simplest approach, where the target feature is replaced with zeros.

Relative transformation operates by computing the difference in target feature values between two adjacent packets. This strategy specifically targets features classified as relative artifacts. While the absolute values of such features are typically initialized based on the TCP connection startup timestamp, their differences encode meaningful semantics. For instance, the difference in TCP sequence numbers reflects packet length, whereas the difference in TCP timestamp options represents packet inter-arrival time.

Random masking provides a more robust occlusion mechanism. We randomize both source and destination IP addresses and ports of the first packet, then adjust those in subsequent packets within the session accordingly. For relative artifacts, we randomize the feature values of the first packet while preserving the relative differences in subsequent packets. For all other features, field values are randomized independently.

% We evaluate model performance using standard metrics to evaluate the classification performance across datasets and models. :
% \begin{itemize}
%     \item \textbf{Accuracy (AC):} Fraction of correctly classified samples.
%     % \item \textbf{Precision (PR):} Number of true positives divided by the sum of true positives and false positives.
%     % \item \textbf{Recall (RC):} Ability of the method to identify all relevant instances (true positives divided by the sum of true positives and false negatives).
%     \item \textbf{F1 Score (F1):} Harmonic mean of precision and recall, weighted for multi-class cases.
% \end{itemize}

% /mnt/ssd1/wtz_nta_dataset/CICMalAnal2017/Adware/Gooligan/06_15_2017-ad-gooligan-fortinet-eda506a6c01c3c7e149ebaebcf929c40.pcap
% one of the dataset's pcap appears to have been cut short in the middle of a packet. So we choose to remove all of the pcap.

\subsection{Detection Experiments}

% \begin{figure*}[htpb]
%     \centering
%     \includegraphics[scale=0.28]{pics/APP1.pdf}
%     \includegraphics[scale=0.28]{pics/APP2.pdf}
%     \caption{Encrypted Application Classification Datasets}
%     \label{fig:app}
% \end{figure*}

% \begin{figure*}[htpb]
%     \centering
%     \includegraphics[scale=0.5]{pics/tsval.pdf}
%     \includegraphics[scale=0.5]{pics/ack.pdf}
%     \includegraphics[scale=0.5]{pics/seq.pdf}
%     \caption{Relative-Value Shortcut Features}
%     \label{fig:app}
% \end{figure*}

% \begin{figure*}[htpb]
%     \centering
%     \includegraphics[scale=0.4]{pics/line_chart.pdf}
%     \caption{AMI Comparison of Relative Artifacts Across Three Tasks}
%     \label{fig:relative}
% \end{figure*}

% \begin{figure}[htbp]
%     \centerline{\includegraphics[scale=0.40]{pics/heatmap_Blues.pdf}}
%     \caption{Average AMI Scores of Key Features Across Different Tasks}
%     \label{fig:heatmap}
% \end{figure}

\begin{figure*}[htbp]
  \centering
  \begin{subfigure}[t]{\textwidth}
    \centering
    \includegraphics[height=7cm, keepaspectratio]{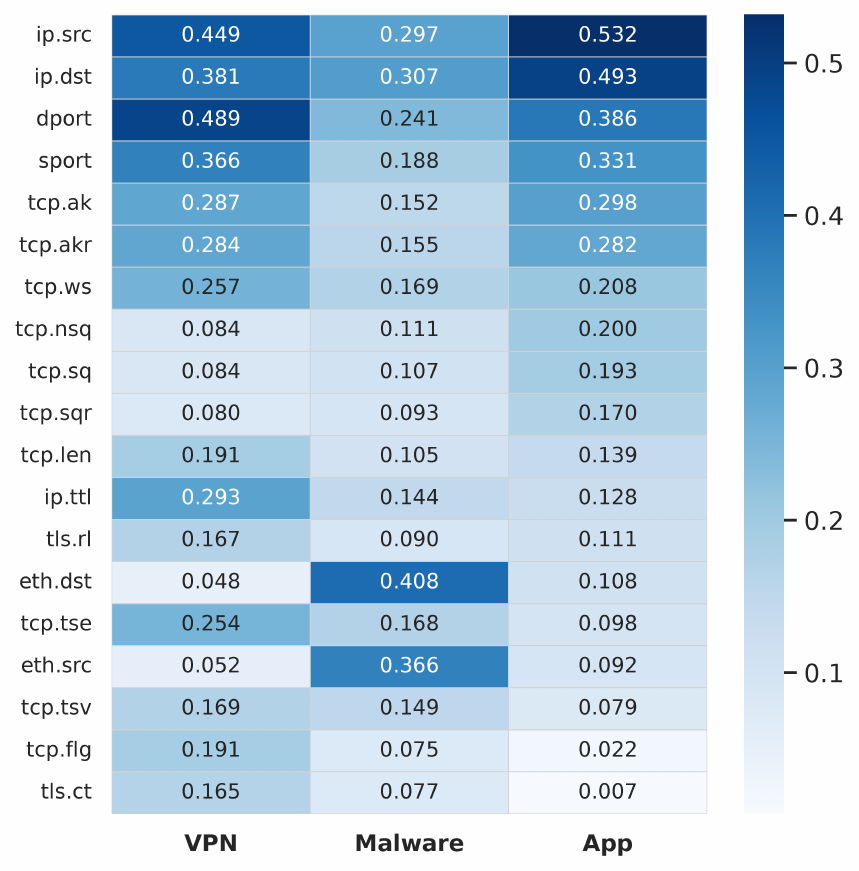}
    \hfill
    \includegraphics[height=7cm, keepaspectratio]{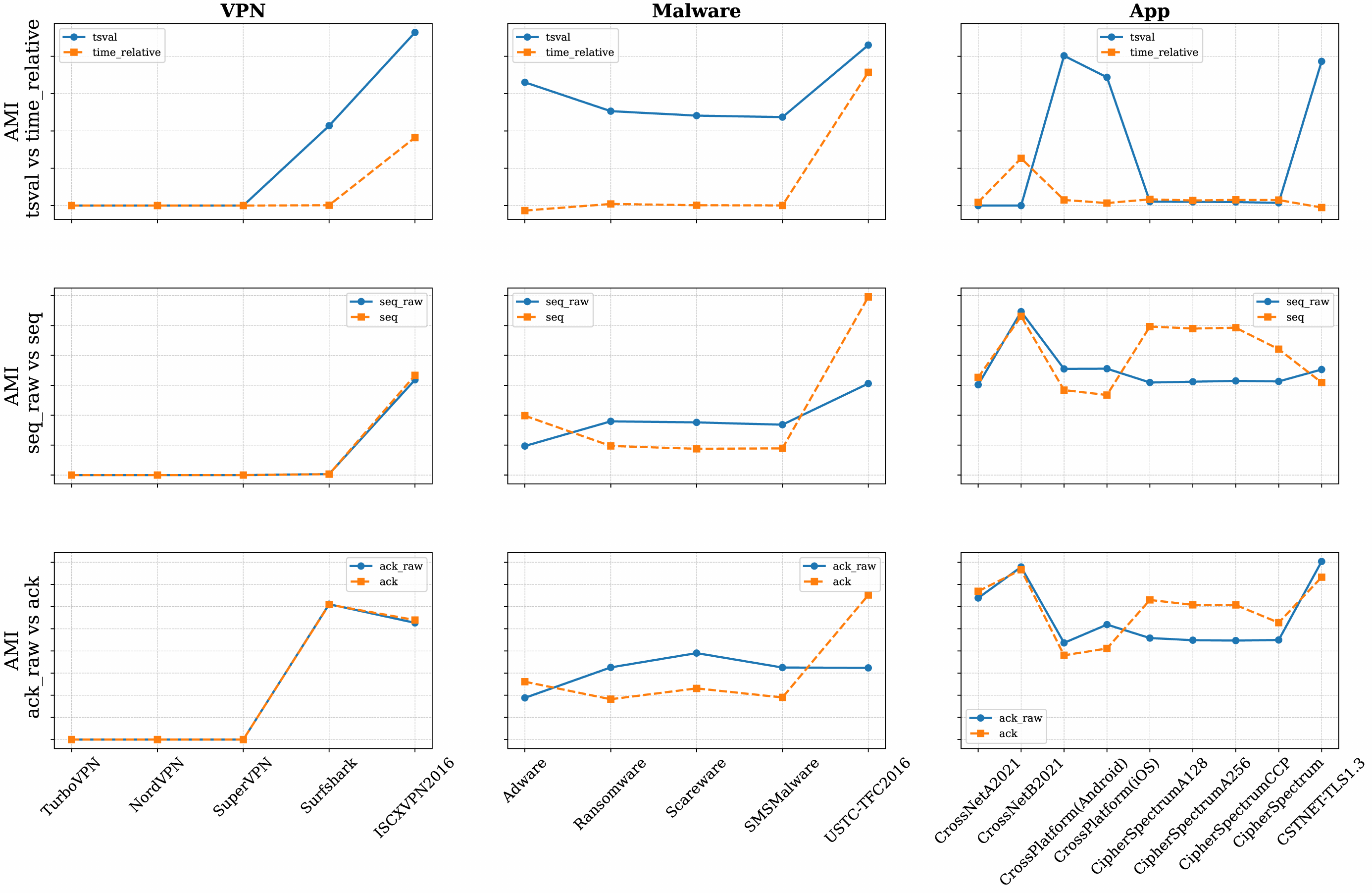}
  \end{subfigure}
  \caption{(L) Average AMI Scores of Key Features Across Tasks; (R) AMI Comparison of Relative Artifacts Across Tasks.}
  \label{fig:side_by_side}
\end{figure*}

We present experimental evidence supporting our shortcut detection framework. Our analysis is organized into four major components: (1) top-k shortcut field selection using AMI, (2) task-specific and cross-task feature behaviors, and (3) granular inspection of different types of suspected shortcuts.

\subsubsection{\textbf{Top-k Selection and Hypothesis}}
We compute Adjusted Mutual Information (AMI) between each field and the class label across three classification tasks: VPN detection, malware detection, and encrypted app classification. Features are then ranked by their AMI scores, and the top-k fields are identified as potentially shortcut-prone. Here we present the AMI top10 ranked features for the VPN detection task, as illustrated in Figure~\ref{fig:vpn}.  
% \textcolor{red}{TODO: summarize the results of Fig. 2}
% Results are shown in figure \ref{fig:vpn} \ref{fig:malware} and \ref{fig:app}.

\subsubsection{\textbf{Task-Specific and Cross-Task Feature Behavior}}
As shown in Fig~\ref{fig:side_by_side}(L), identifier fields such as \textit{ip.src}, \textit{ip.dst}, \textit{srcport} and \textit{dstport} consistently appear among the top-ranked fields across all tasks (VPN, malware, and app classification), 
% \textcolor{red}{TODO: why these featrues are data-leakage identifiers? where is the categorization?} 
implying a generalized shortcut tendency. These fields, while highly discriminative, often reflect dataset artifacts rather than meaningful semantics.

While demonstrating notable feature stability across tasks, our analysis reveals distinct task-specific field preferences:
\begin{itemize}
    \item VPN Traffic Classification: Relies heavily on IP address and port-based features.
    \item Malware Traffic Classification: Leverages MAC and timing-level features such as \textit{eth.dst} and \textit{tcp.window\_size}.
    \item Encrypted Application Classification: Exhibits mixed-layer reliance, including TLS-layer metadata, sequence numbers, and TCP timestamp options.
\end{itemize}

These findings support the existence of both task-shared and task-specific shortcut fields and reinforce the necessity of careful feature selection and mitigation strategies to ensure robust encrypted traffic classification.

\subsubsection{\textbf{Granular Feature Analysis}}

We further partition AMI-ranked features into three categories(shown in Table~\ref{tab:feature_categorization}) for detailed analysis according to further experimental validations.

\textbf{Data-leakage identifiers}.
Fields such as TCP Server Name Indication (SNI) and other Strong Identification Indicators (SII), including source/destination IP addresses and ports, serve as strong signals of potential overfitting, as they are often directly used for labeling or exhibit high correlation with labels. Leveraging domain knowledge, these fields can be identified and are typically removed in fair evaluation settings.

\textbf{Relative artifacts}.
Due to their large bit widths, TCP-level fields such as timestamps, sequence numbers, and acknowledgment numbers within the same session often exhibit constant high-order bits, making them potential shortcut features. To mitigate shortcuts arising from these high-order bits, we apply relative transformations to these fields (e.g., converting \textit{tsval} to the difference \textit{tsval}--\textit{tsecr}, or \textit{seq\_raw} to normalized \textit{seq}). We then compute the AMI of both absolute and relative versions across datasets and plot them for visual comparison.

Specifically, we evaluate three field pairs: (\textit{tsval} vs. \textit{time\_relative}), (\textit{seq\_raw} vs. \textit{seq}), and (\textit{ack\_raw} vs. \textit{ack}). These pairs reflect a transition from raw values to semantics-aware or relative measurements. Figure~\ref{fig:side_by_side}(R) presents a row-wise comparison across three traffic classification scenarios (VPN, Malware, and Encrypted Application), with each column representing different datasets within the same task, and each row corresponding to one of the three field pairs. The vertical axis denotes the AMI score, representing feature-task alignment.

\begin{itemize}
    \item \textbf{Malware traffic classification} exhibits the most significant reduction in AMI after relative transformation. Particularly for timestamp-related fields (\textit{tsval} vs. \textit{time\_relative}), the raw field consistently shows strong task alignment, while its relative counterpart demonstrates reduced utility. This suggests that malware traffic families may encode fixed or correlated timing patterns exploitable by the classifier.
    
    \item \textbf{VPN traffic classification} displays the lowest AMI drop across all three field pairs. The relative and absolute versions of sequence and acknowledgement fields show comparable AMI values, indicating that these fields do not serve as strong shortcuts. VPN tools might exhibit more consistent behavior across instances, limiting the impact of alignment-based artifacts.
    
    \item In \textbf{Encrypted Application classification}, relative timestamp fields (\textit{time\_relative}) show lower AMI than their absolute counterparts (\textit{tsval}), but the drop is less severe than in Malware. Notably, this difference is magnified in Cross-Platform datasets (e.g., CrossPlatform-Android vs. CrossPlatform-iOS), suggesting that OS-level stack behaviors introduce shortcut potentials in timestamp fields.
\end{itemize}

% This comparison highlights that absolute field values—especially timestamps—may encode environment-specific or implementation-specific artifacts that can mislead the classifier. Converting to relative representations not only reduces shortcut reliance but also improves generalization across domains.
This comparison reveals that absolute field values may encode environment-specific or implementation-specific artifacts that adversely affect classifier generalization. Converting to relative representations not only mitigates reliance on these artifacts but also enhances cross-domain generalization capability.

\textbf{Task-Agnostic fields}.
Fields such as IP TTL, IP checksum, TCP window size, and TCP checksum carry limited relevance to NTC tasks. For these fields, we inspect their statistical distributions and environmental sensitivity. Among these, we select \textit{TCP Window Size} as a representative example to illustrate our analysis procedure, due to its known dependence on bandwidth and congestion dynamics.

Defined as the amount of data a sender is willing to receive, TCP Window Size is known to vary with bandwidth and congestion conditions. We conduct a comparison using two datasets: CrossNet-A and CrossNet-B.
\begin{itemize}
    \item \textbf{Scenario A} was collected under a stable 100 Mbps environment with low latency and no artificial delay or jitter.
    \item \textbf{Scenario B} comes from a 10 Mbps network with a random packet loss rate (2.5\% to 5\%) and approximately 200 ms of delay due to unstable links.
\end{itemize}

\begin{figure}[htbp]
    \centerline{\includegraphics[scale=0.22]{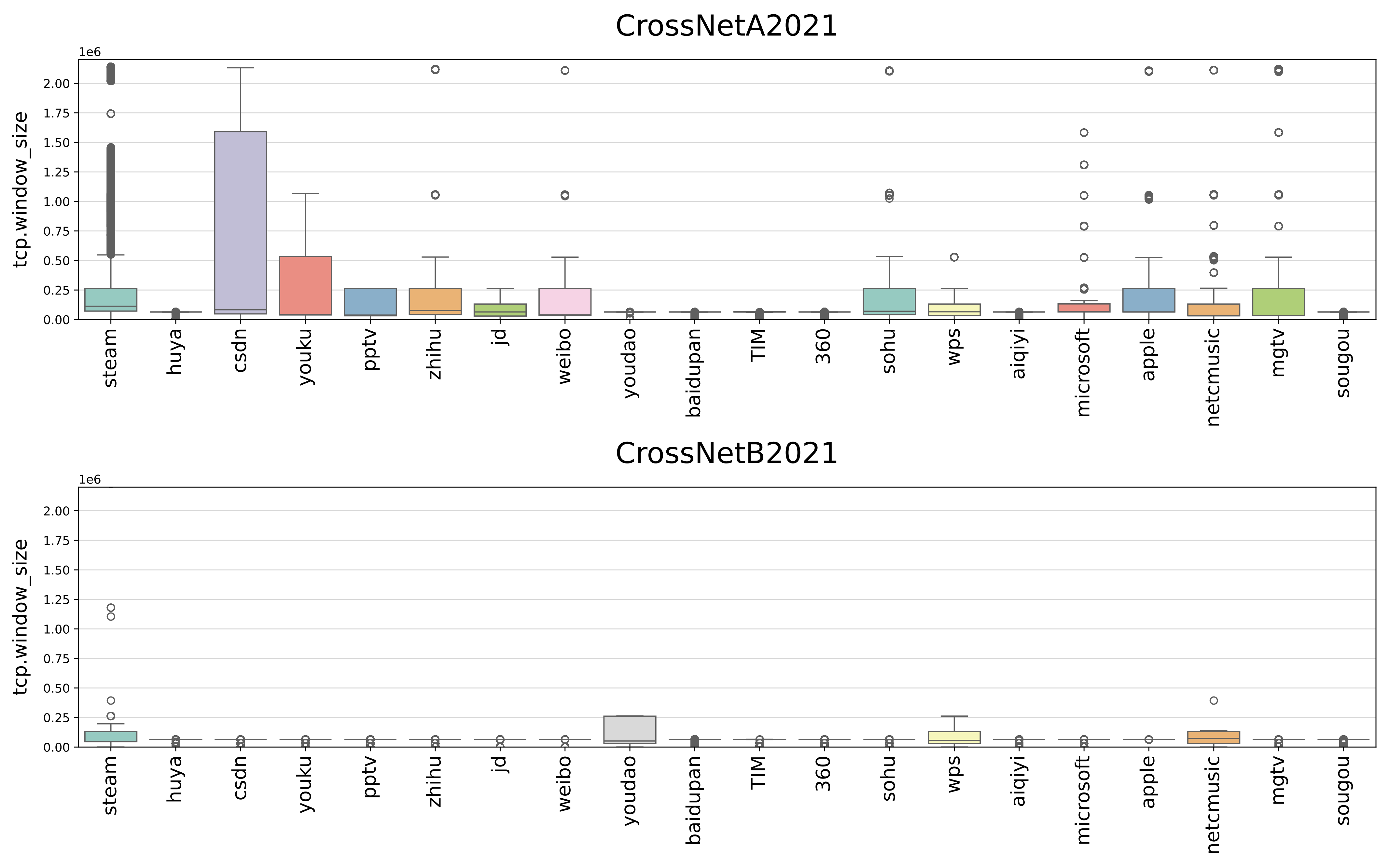}}
    \caption{TCP Window Size Distributions Across Different Network Quality Conditions}
    \label{fig:window}
\end{figure}

The distributions of TCP Window Size for different applications in both datasets are shown in Figure~\ref{fig:window}. We observe significant distributional differences for the same application type under different network conditions. The window size distribution in CrossNet-B is more concentrated, while CrossNet-A shows a more dispersed distribution, indicating that network conditions significantly impact TCP Window Size.

To further quantify the correlation between TCP Window Size and network quality, we employ KDE to estimate the probability density distributions of the field in both datasets and compute KL divergence to measure distributional differences. For intuitive comparison, we also calculate KL divergence for more stable fields like TCP length, TCP flags, and IP length. The KL divergence of TCP Window Size (2.36) is significantly higher than others (e.g., TCP length (0.06), TCP flags (0.53), IP length (0.05)), validating its sensitivity to network conditions and potential role as an unstable feature.
% As shown in Table~\ref{tab:kl_divergence}, 

We evaluate other fields such as Checksum and IP TTL using similar methods. While some fields exhibit mild sensitivity due to routing behavior differences, others like Checksum are primarily determined by payload content and packet structure, showing less environmental variance. Combining domain knowledge, IP Checksum is used to verify the integrity of packet header bytes, where any change in header bytes causes Checksum value changes, making it semantically unrelated to network traffic classification. IP TTL reflects the number of hops during packet transmission, typically preventing infinite loops in networks. Although TTL values may vary with network topology or routing policies, they generally do not directly impact traffic classification outcomes.

% \begin{table}[htbp]
% \centering
% \renewcommand{\arraystretch}{1.3}
% \setlength{\tabcolsep}{8pt}
% \caption{Categorization of Shortcut-Prone Feature Fields}
% \label{tab:shortcut-categories}
% \begin{tabular}{p{4.5cm} p{7.5cm}}
% \toprule
% \textbf{Feature Category} & \textbf{Representative Fields} \\
% \midrule
% \textbf{Data-Leakage Identifiers} 
% & 
% \texttt{ip.src}, \texttt{ip.dst}, \texttt{tcp.srcport}, \texttt{tcp.dstport}; \newline
% \texttt{tls.handshake.extensions\_server\_name} \\
% \midrule
% \textbf{Relative Artifacts} 
% & 
% \texttt{tcp.seq}, \texttt{tcp.ack}, \texttt{tcp.options.timestamp.tsval}, \texttt{tcp.options.timestamp.tsecr} \\
% \midrule
% \textbf{Task-Agnostic Fields} 
% & 
% \texttt{tcp.window\_size\_value}, \texttt{tcp.checksum}, \texttt{ip.checksum}, \texttt{ip.ttl} \\
% \bottomrule
% \end{tabular}
% \end{table}

\begin{table}[htbp]
\centering
\caption{Potential Shortcut Feature}
\label{tab:feature_categorization}
\begin{tabular}{>{\raggedright\arraybackslash}p{2.9cm}|>{\raggedright\arraybackslash}p{5.1cm}}
\toprule
\textbf{Category} & \textbf{Representative Features} \\
\midrule
Data-Leakage Identifiers &
SII(Src IP, Dst IP, Src Port, Dst Port), SNI \\
\addlinespace
Relative Artifacts &
Seq Num, Ack Num, Timestamp Option\\
\addlinespace
Task-Agnostic Fields &
Window Size, Checksum, TTL \\
\bottomrule
\end{tabular}
\end{table}

\subsection{Mitigation Evaluation}

\newcommand{\betweenTinyAndScript}{\fontsize{6.5pt}{7.5pt}\selectfont}
\begin{table*}[htpb]
    \centering
    \betweenTinyAndScript
    \caption{Classification Accuracies on Different Shortcut Feature Mitigation Strategies}
    \begin{threeparttable}
    \begin{tabular}{c|c|cc|cc|cc|cc|cc|cc}
    \toprule
    \multirow{2}{*}{Shortcut}& \multirow{2}{*}{Mitigation Strategy}& \multicolumn{2}{c|}{CrossNet2021\cite{li2022robust}} & \multicolumn{2}{c|}{CSTNET-TLS1.3\cite{lin2022bert}} & \multicolumn{2}{c|}{ISCXVPN2016\cite{gil2016characterization}} &\multicolumn{2}{c|}{SurfsharkVPN\cite{oh2023appsniffer}} & \multicolumn{2}{c|}{USTC-TFC2016\cite{wang2017malware}} & \multicolumn{2}{c}{Ransomware\cite{HabibiLashkari2018TowardDA}} \\
    \cmidrule(lr){3-4} \cmidrule(lr){5-6} \cmidrule(lr){7-8} \cmidrule(lr){9-10} \cmidrule(lr){11-12} \cmidrule(lr){13-14}
    & & NM & DT & NM & DT & NM & DT & NM & DT & NM & DT & NM & DT \\
    \midrule
    None & Full Feature & 0.9472 & 0.8936 & 0.9891 & 0.9703 & 0.8847 & 0.9132 & 0.9619 & 0.9661 & 0.9799 & 0.9810 & 0.3905 & 0.3810 \\ \midrule
    \multirow{4}{*}{DL} & Zero SII & \textcolor{SkyBlue}{0.9213} & \textcolor{SkyBlue}{0.8786} & \textcolor{SkyBlue}{0.9733} & \textcolor{SkyBlue}{0.9564} & \textcolor{SkyBlue}{0.8314} & \textcolor{SkyBlue}{0.8590} & \textcolor{SkyBlue}{0.9405} & \textcolor{SkyBlue}{0.9579} & \textcolor{SkyBlue}{0.9656} & \textcolor{SkyBlue}{0.9654} & \textcolor{SkyBlue}{0.2871} & \textcolor{SkyBlue}{0.2718} \\
    & Zero SNI & \textcolor{SkyBlue}{0.9294} & \textcolor{SkyBlue}{0.8794} & \textcolor{SkyBlue}{0.9862} & \textcolor{SkyBlue}{0.9657} & \textcolor{Purple}{0.8858} & \textcolor{Purple}{0.9143} & \textcolor{Purple}{0.9776} & \textcolor{SkyBlue}{0.9598} & \textcolor{Purple}{0.9840} & \textcolor{SkyBlue}{0.9797} & \textcolor{SkyBlue}{0.3785} & \textcolor{SkyBlue}{0.3696} \\
    & Random SII & \textcolor{SkyBlue}{0.9164} & \textcolor{SkyBlue}{0.8561} & \textcolor{SkyBlue}{0.9812} & \textcolor{SkyBlue}{0.9576} & \textcolor{SkyBlue}{0.7821} & \textcolor{SkyBlue}{0.8339} & \textcolor{SkyBlue}{0.9155} & \textcolor{SkyBlue}{0.9440} & \textcolor{SkyBlue}{0.9732} & \textcolor{SkyBlue}{0.9680} & \textcolor{SkyBlue}{0.2552} & \textcolor{SkyBlue}{0.2574} \\
    & Random SNI & \textcolor{SkyBlue}{0.9142} & \textcolor{SkyBlue}{0.8752} & \textcolor{SkyBlue}{0.9877} & \textcolor{SkyBlue}{0.9635} & \textcolor{Purple}{0.8982} & \textcolor{Purple}{0.9145} & \textcolor{SkyBlue}{0.9141} & \textcolor{SkyBlue}{0.9423} & \textcolor{SkyBlue}{0.9796} & \textcolor{SkyBlue}{0.9800} & \textcolor{Purple}{0.4139} & \textcolor{Purple}{0.3983} \\ \midrule
    \multirow{6}{*}{RA} & Zero TCP Timestamp & \textcolor{SkyBlue}{0.9333} & \textcolor{SkyBlue}{0.8830} & \textcolor{SkyBlue}{0.9839} & \textcolor{SkyBlue}{0.9623} & \textcolor{SkyBlue}{0.8708} & \textcolor{SkyBlue}{0.8771} & \textcolor{SkyBlue}{0.9585} & \textcolor{SkyBlue}{0.9474} & \textcolor{Purple}{0.9840} & \textcolor{Purple}{0.9820} & \textcolor{Purple}{0.4075} & \textcolor{SkyBlue}{0.3495} \\
    & Zero SEQ/ACK No & \textcolor{SkyBlue}{0.9358} & \textcolor{SkyBlue}{0.8922} & \textcolor{SkyBlue}{0.9829} & \textcolor{SkyBlue}{0.9672} & \textcolor{SkyBlue}{0.8835} & \textcolor{SkyBlue}{0.9096} & \textcolor{SkyBlue}{0.8566} & \textcolor{SkyBlue}{0.9523} & \textcolor{Purple}{0.9864} & \textcolor{Purple}{0.9840} & \textcolor{Purple}{0.4136} & \textcolor{Purple}{0.3867} \\
    & Relative TCP Timestamp & \textcolor{SkyBlue}{0.9416} & \textcolor{Purple}{0.8984} & \textcolor{SkyBlue}{0.9857} & \textcolor{SkyBlue}{0.9617} & \textcolor{SkyBlue}{0.8697} & \textcolor{SkyBlue}{0.8835} & \textcolor{SkyBlue}{0.9373} & \textcolor{SkyBlue}{0.9508} & \textcolor{Purple}{0.9813} & \textcolor{SkyBlue}{0.9787} & \textcolor{SkyBlue}{0.3877} & \textcolor{SkyBlue}{0.3732} \\
    & Relative SEQ/ACK No & \textcolor{SkyBlue}{0.9248} & \textcolor{SkyBlue}{0.8857} & \textcolor{SkyBlue}{0.9798} & \textcolor{SkyBlue}{0.9692} & \textcolor{SkyBlue}{0.8786} & \textcolor{SkyBlue}{0.9032} & \textcolor{SkyBlue}{0.8701} & \textcolor{SkyBlue}{0.9615} & \textcolor{Purple}{0.9833} & \textcolor{Purple}{0.9813} & \textcolor{SkyBlue}{0.3882} & \textcolor{SkyBlue}{0.3801} \\
    & Random TCP Timestamp & \textcolor{SkyBlue}{0.9394} & \textcolor{SkyBlue}{0.8923} & \textcolor{SkyBlue}{0.9838} & \textcolor{Purple}{0.9727} & \textcolor{SkyBlue}{0.8564} & \textcolor{SkyBlue}{0.8753} & \textcolor{Purple}{0.9640} & \textcolor{SkyBlue}{0.9527} & \textcolor{Purple}{0.9826} & \textcolor{SkyBlue}{0.9800} & \textcolor{Purple}{0.4035} & \textcolor{SkyBlue}{0.3641} \\
    & Random SEQ/ACK No & \textcolor{SkyBlue}{0.9250} & \textcolor{SkyBlue}{0.8787} & \textcolor{SkyBlue}{0.9880} & \textcolor{Purple}{0.9713} & \textcolor{Purple}{0.8890} & \textcolor{SkyBlue}{0.9012} & \textcolor{SkyBlue}{0.8527} & \textcolor{SkyBlue}{0.9547} & \textcolor{Purple}{0.9847} & \textcolor{SkyBlue}{0.9803} & \textcolor{SkyBlue}{0.3632} & \textcolor{Purple}{0.3888} \\ \midrule
    \multirow{8}{*}{TA} & Zero IP TTL & \textcolor{SkyBlue}{0.9420} & \textcolor{SkyBlue}{0.8803} & \textcolor{SkyBlue}{0.9857} & \textcolor{SkyBlue}{0.9701} & \textcolor{SkyBlue}{0.8705} & \textcolor{Purple}{0.9163} & \textcolor{SkyBlue}{0.9506} & \textcolor{SkyBlue}{0.9539} & \textcolor{SkyBlue}{0.9786} & \textcolor{SkyBlue}{0.9784} & \textcolor{Purple}{0.3971} & \textcolor{Purple}{0.3965} \\
    & Zero TCP Window & \textcolor{SkyBlue}{0.9281} & \textcolor{SkyBlue}{0.8667} & \textcolor{SkyBlue}{0.9880} & \textcolor{SkyBlue}{0.9481} & \textcolor{Purple}{0.8898} & \textcolor{SkyBlue}{0.9077} & \textcolor{SkyBlue}{0.9598} & \textcolor{SkyBlue}{0.9456} & \textcolor{Purple}{0.9863} & \textcolor{SkyBlue}{0.9756} & \textcolor{Purple}{0.4057} & \textcolor{black}{0.3810} \\
    & Zero IP Checksum & \textcolor{SkyBlue}{0.9412} & \textcolor{SkyBlue}{0.8900} & \textcolor{SkyBlue}{0.9808} & \textcolor{SkyBlue}{0.9695} & \textcolor{SkyBlue}{0.8778} & \textcolor{SkyBlue}{0.9072} & \textcolor{SkyBlue}{0.9434} & \textcolor{SkyBlue}{0.9360} & \textcolor{Purple}{0.9830} & \textcolor{Purple}{0.9840} & \textcolor{Purple}{0.3937} & \textcolor{Purple}{0.3903} \\
    & Zero TCP/UDP Checksum & \textcolor{SkyBlue}{0.9406} & \textcolor{SkyBlue}{0.8910} & \textcolor{SkyBlue}{0.9859} & \textcolor{SkyBlue}{0.9641} & \textcolor{SkyBlue}{0.8844} & \textcolor{Purple}{0.9201} & \textcolor{SkyBlue}{0.9431} & \textcolor{SkyBlue}{0.9544} & \textcolor{Purple}{0.9829} & \textcolor{SkyBlue}{0.9776} & \textcolor{SkyBlue}{0.3871} & \textcolor{Purple}{0.3988} \\
    & Random IP TTL & \textcolor{SkyBlue}{0.9197} & \textcolor{SkyBlue}{0.8606} & \textcolor{Purple}{0.9898} & \textcolor{SkyBlue}{0.9682} & \textcolor{SkyBlue}{0.8691} & \textcolor{SkyBlue}{0.9019} & \textcolor{SkyBlue}{0.9051} & \textcolor{SkyBlue}{0.9486} & \textcolor{SkyBlue}{0.9797} & \textcolor{black}{0.9810} & \textcolor{SkyBlue}{0.3611} & \textcolor{SkyBlue}{0.3664} \\
    & Random TCP Window & \textcolor{SkyBlue}{0.9201} & \textcolor{SkyBlue}{0.8652} & \textcolor{SkyBlue}{0.9857} & \textcolor{SkyBlue}{0.9562} & \textcolor{SkyBlue}{0.8825} & \textcolor{SkyBlue}{0.8996} & \textcolor{SkyBlue}{0.9231} & \textcolor{SkyBlue}{0.9346} & \textcolor{Purple}{0.9808} & \textcolor{black}{0.9810} & \textcolor{Purple}{0.3984} & \textcolor{SkyBlue}{0.3739} \\
    & Random IP Checksum & \textcolor{SkyBlue}{0.9212} & \textcolor{SkyBlue}{0.8745} & \textcolor{SkyBlue}{0.9827} & \textcolor{SkyBlue}{0.9645} & \textcolor{SkyBlue}{0.8716} & \textcolor{SkyBlue}{0.9000} & \textcolor{SkyBlue}{0.9502} & \textcolor{SkyBlue}{0.9566} & \textcolor{Purple}{0.9803} & \textcolor{SkyBlue}{0.9766} & \textcolor{Purple}{0.3919} & \textcolor{Purple}{0.3961} \\
    & Random TCP/UDP Checksum & \textcolor{SkyBlue}{0.9368} & \textcolor{SkyBlue}{0.8759} & \textcolor{SkyBlue}{0.9801} & \textcolor{SkyBlue}{0.9650} & \textcolor{Purple}{0.8912} & \textcolor{SkyBlue}{0.9100} & \textcolor{SkyBlue}{0.9366} & \textcolor{SkyBlue}{0.9544} & \textcolor{Purple}{0.9821} & \textcolor{SkyBlue}{0.9779} & \textcolor{Purple}{0.4159} & \textcolor{Purple}{0.3828} \\
    \bottomrule
    \end{tabular}
    \begin{tablenotes}
        \item[1] NM: NetMamba, DT: Decision Tree.
        \item[2] DL: Data Leakage, RA: Relative Artifacts, TA: Task Agnostic.
        \item[3] We highlight the metric higher than the full feature in \textcolor{Purple}{purple} and lower in \textcolor{SkyBlue}{blue}.
    \end{tablenotes}
    \end{threeparttable}
    \label{tb:acc-occlusion}
\end{table*}

% \textcolor{blue}{CSTNET-TLS1.3 wo/ SNI:} SNIs are already removed in the released datasets, thus no accuracy drops are observed.

% \textcolor{blue}{USTC-TFC wo/ SNI:} Accuracies achieved improvement, it is an interesting phenomenon, maybe we can count and analyse the SNI distribution within USTC-TFC2016.

% \textcolor{red}{TODO:} Add more datasets and models.
% For datasets, consider another VPN (which one?) and Malware (CIC-AndMal2017); for models, consider YaTC (or NetTrans) if in time.

We further examine and mitigate the influence of shortcut features on NTC models, with the corresponding evaluation results presented in Table~\ref{tb:acc-occlusion}. 
Prior studies~\cite{guthula2023netfound, oh2023appsniffer, wickramasinghe2025sok} have reported that the classification accuracy of NTC models typically decreases when shortcut features are mitigated in the input data.
However, as shown in Table~\ref{tb:acc-occlusion}, we unexpectedly observe that both NTC models achieve accuracy improvements under several occlusion settings. Our analysis is as follows:

% \subsubsection{Data-leakage identifiers and task-agnostic fields are spurious correlations.}
% By setting shortcut features belonging to data-leakage identifiers and task-agnostic fields to zero, we found NetMamba's accuracies dropped on most datasets. This verifies the NetMamba relies on these unintended features to identify network traffic.

% \subsubsection{Relative artifacts contain useful network dynamics.}
% Compared to directly zero padding, NetMamba's performance increases on most settings when applying relative transformation to shortcut features belonging to relative artifacts. 
% This further emphasizes the value of network dynamics (i.e., packet length and inter-arrival times) for traffic classification.

% \subsubsection{Shortcut feature's effect differs across datasets}
% Once strong shortcut features, such as SII, TTL, and IP Checksum, are removed, NetMamba's performance drops greatly on datasets like CrossNet2021 and CSTNET-TLS1.3, but increases on SurfsharkVPN. This indicates us to find task-specific or scenario-specific shortcuts when developping traffic classification models.  

\subsubsection{NTC models exhibit susceptibility to shortcut features}
With the exception of USTC-TFC2016 and Ransomware, both NTC models experience accuracy degradation under most occlusion settings for the remaining four datasets. This suggests that, prior to occlusion, the models partially rely on shortcut features for classification.

\subsubsection{Data leakage constitutes a stable shortcut category}
Across all six datasets, removing SII or SNI consistently results in noticeable accuracy drops. In contrast, removing relative artifacts or task-agnostic fields produces less consistent effects. This indicates that strong identifiers serve as critical shortcuts for NTC models across these three task types.

\subsubsection{Random masking introduces additional noise} Compared with zero padding, random masking injects extra noise into the traffic features, causing larger average performance degradation. However, spatial or temporal patterns introduced by relative transformation have relatively minor effects compared to the other occlusion strategies.

\section{Discussion}
\label{sec:result}

Shortcut features in encrypted network traffic classification are not universally harmful nor easily dismissed. Instead, their presence and influence are deeply tied to dataset-specific properties, labeling strategies, and operational contexts. We introduce \textit{BiasSeeker}, a hybrid detection framework that combines statistical analysis with domain-informed validation to identify shortcut-prone features. Beyond this, we argue that feature selection must become a deliberate preliminary step in model design, grounded in thorough assessment of data characteristics and deployment constraints.

\subsection{The BiasSeeker Framework: A Hybrid Approach}
Our goal is to design a model-agnostic framework generalizable across architectures. BiasSeeker implements a three-stage pipeline: (1) data-driven feature ranking, (2) candidate filtering via domain knowledge, and (3) category-specific validation strategies. This structured process integrates data signals with domain reasoning to ensure systematic shortcut identification, moving beyond purely manual approaches while delivering clear, actionable outcomes.

\subsubsection{\textbf{Data-Driven Feature Prioritization via AMI}}
We employ Adjusted Mutual Information (AMI) as a robust, interpretable, and low-cost statistical tool to prioritize features by capturing their first-order correlations with labels. This approach is justified by three key advantages:
\begin{itemize}
    \item First, AMI serves as a transparent, model-agnostic metric that operates independently of specific models or training dynamics.
    \item Second, its ability to detect statistical associations is particularly critical in encrypted traffic contexts where feature semantics are often unknown.
    \item Crucially, this method systematically narrows the candidate space of potential shortcut features, compressing the manual inspection scope from hundreds of raw features to a limited candidate set.
\end{itemize} 

By leveraging AMI's statistical screening instead of relying on prior protocol semantics, we significantly reduce dependency on expert knowledge while maintaining detection efficacy. This strategy proves indispensable for high-dimensional traffic datasets where exhaustive manual feature examination is computationally infeasible.

\subsubsection{\textbf{Domain-Enhanced Feature Interpretation Pipeline}}
Once highly correlated features are identified, we incorporate domain knowledge to interpret their semantics and assess whether they are likely to be spurious or task-relevant. We then categorize these features into intuitive groups, each with an associated validation strategy tailored to its nature.

We acknowledge that the final decision still requires human inspection. However, this is a significant improvement over fully manual processes, which require inspecting hundreds of raw features. BiasSeeker acts as a taxonomy system, highlighting suspicious signals that are more likely to encode dataset-specific bias.

Building an automatic shortcut detector is a challenging open problem. We hope our framework provides a structured first step, and we are actively exploring causal analysis and representation-based criteria for more automated shortcut diagnosis in future work.

\textit{\textbf{Takeaway 1}: BiasSeeker effectively narrows down potential shortcut features by combining statistical screening, significantly reducing manual inspection effort, and provides a first step toward automated shortcut detection in high-dimensional traffic datasets.}

\subsection{Shortcut Features Are Contextual, Not Universal}

Shortcut features should not be treated as inherently flawed nor universally detrimental. Our study underscores the necessity of dataset-specific, context-aware analysis when identifying and mitigating shortcut behaviors. Across our evaluations, we observe that certain features may act as strong shortcut indicators in some datasets, yet serve valid semantic or operational roles in others. For instance, SNI may be a valid signal in app identification, but fragile in malware detection due to dynamic evasion. These discrepancies arise from variations in traffic composition, network conditions, labeling strategies, and deployment environments.

Importantly, we do not advocate for a blanket exclusion of such fields. In certain scenarios—e.g., controlled industrial deployments or static topologies—these features may yield practical benefits despite limited generalizability. However, their inclusion must be preceded by critical examination of environmental coupling, transferability risks, and labeling entanglement.

Our findings are not intended to be prescriptive or exhaustive but rather to raise awareness. Shortcut detection must be grounded in the specific goals, data properties, and deployment constraints of each use case. Feature selection should thus be an intentional and scenario-sensitive step prior to model training or benchmarking.

\textit{\textbf{Takeaway 2}: Shortcut features are not inherently harmful; their impact is highly dataset- and context-dependent. Feature selection should be a scenario-sensitive process that considers environmental coupling and transferability before model training.}

\subsection{Semantic Learning vs. Shortcut Reliance in Encrypted NTC}

Our findings prompt a fundamental reconsideration: are current traffic classification models truly capturing the semantic structure of encrypted flows, or are they instead overfitting to superficial, dataset-specific artifacts? As network protocols, application behaviors, and operational environments continue to evolve, models built on brittle correlations—such as fixed IP mappings or length-based heuristics—face rapid obsolescence.

This prompts a rethinking of design goals in encrypted NTC. To foster robust and future-proof classifiers, we advocate for a shift from accuracy-centric optimization to resilience-oriented modeling, and argue that future research should prioritize the following directions:
\begin{itemize}
    \item \textbf{Learning semantic, protocol-invariant representations} that remain stable across evolving network conditions and deployment scenarios.
    \item \textbf{Systematic detection and mitigation of shortcut features}, especially those arising from collection artifacts or labeling leakage.
    \item \textbf{Development of realistic and diverse benchmarks} that reflect the operational heterogeneity of real-world environments, rather than static or synthetic setups.
\end{itemize}

By shifting the focus from short-term accuracy to long-term resilience, the community can better align network traffic classification research with practical deployment needs and evolving security challenges.

\textit{\textbf{Takeaway 3}: Many models rely on dataset-specific shortcuts rather than true encrypted-flow semantics. Future work should emphasize resilient, semantic representations and realistic benchmarks over short-term accuracy.}

\section{Conclusion}
\label{sec:conclusion}

We presented \textbf{BiasSeeker}, a semi-automated framework that combines model-agnostic analysis with data-centric insights to address shortcut learning in encrypted network traffic classification (NTC). Recognizing the limitations of existing approaches—which often depend on expert heuristics or are tightly coupled with specific model architectures—BiasSeeker provides a principled methodology for examining shortcut learning risks inherent in raw traffic data.

Our framework introduces a principled, taxonomy-based understanding of shortcut behaviors, along with category-specific mitigation strategies. Extensive evaluations across 19 public datasets spanning three tasks, demonstrate BiasSeeker’s effectiveness in uncovering hidden biases, and enhancing transferability under distribution shifts.

Importantly, BiasSeeker is not intended to prescribe universal solutions, but to promote awareness and provide actionable tools for shortcut diagnosis. We highlight the need for intentional, context-aware feature selection as a critical step before model design and benchmarking.

\bibliographystyle{IEEEtran}
\bibliography{reference}

\begin{thebibliography}{10}
\providecommand{\url}[1]{#1}
\csname url@samestyle\endcsname
\providecommand{\newblock}{\relax}
\providecommand{\bibinfo}[2]{#2}
\providecommand{\BIBentrySTDinterwordspacing}{\spaceskip=0pt\relax}
\providecommand{\BIBentryALTinterwordstretchfactor}{4}
\providecommand{\BIBentryALTinterwordspacing}{\spaceskip=\fontdimen2\font plus
\BIBentryALTinterwordstretchfactor\fontdimen3\font minus \fontdimen4\font\relax}
\providecommand{\BIBforeignlanguage}[2]{{%
\expandafter\ifx\csname l@#1\endcsname\relax
\typeout{** WARNING: IEEEtran.bst: No hyphenation pattern has been}%
\typeout{** loaded for the language `#1'. Using the pattern for}%
\typeout{** the default language instead.}%
\else
\language=\csname l@#1\endcsname
\fi
#2}}
\providecommand{\BIBdecl}{\relax}
\BIBdecl

\bibitem{zhang2014robust}
J.~Zhang, X.~Chen, Y.~Xiang, W.~Zhou, and J.~Wu, ``Robust network traffic classification,'' \emph{IEEE/ACM transactions on networking}, vol.~23, no.~4, pp. 1257--1270, 2014.

\bibitem{mirsky2018kitsune}
Y.~Mirsky, T.~Doitshman, Y.~Elovici, and A.~Shabtai, ``Kitsune: an ensemble of autoencoders for online network intrusion detection,'' \emph{arXiv preprint arXiv:1802.09089}, 2018.

\bibitem{liu2019fs}
C.~Liu, L.~He, G.~Xiong, Z.~Cao, and Z.~Li, ``Fs-net: A flow sequence network for encrypted traffic classification,'' in \emph{IEEE INFOCOM 2019-IEEE Conference On Computer Communications}.\hskip 1em plus 0.5em minus 0.4em\relax IEEE, 2019, pp. 1171--1179.

\bibitem{piet2023ggfast}
J.~Piet, D.~Nwoji, and V.~Paxson, ``Ggfast: Automating generation of flexible network traffic classifiers,'' in \emph{Proceedings of the ACM SIGCOMM 2023 Conference}, 2023, pp. 850--866.

\bibitem{xue2024fingerprinting}
D.~Xue, M.~Kallitsis, A.~Houmansadr, and R.~Ensafi, ``Fingerprinting obfuscated proxy traffic with encapsulated $\{$TLS$\}$ handshakes,'' in \emph{33rd USENIX Security Symposium (USENIX Security 24)}, 2024, pp. 2689--2706.

\bibitem{lin2022bert}
X.~Lin, G.~Xiong, G.~Gou, Z.~Li, J.~Shi, and J.~Yu, ``Et-bert: A contextualized datagram representation with pre-training transformers for encrypted traffic classification,'' in \emph{Proceedings of the ACM Web Conference 2022}, 2022, pp. 633--642.

\bibitem{zhao2023yet}
R.~Zhao, M.~Zhan, X.~Deng, Y.~Wang, Y.~Wang, G.~Gui, and Z.~Xue, ``Yet another traffic classifier: A masked autoencoder based traffic transformer with multi-level flow representation,'' in \emph{Proceedings of the AAAI Conference on Artificial Intelligence}, vol.~37, no.~4, 2023, pp. 5420--5427.

\bibitem{wang2024netmamba}
T.~Wang, X.~Xie, W.~Wang, C.~Wang, Y.~Zhao, and Y.~Cui, ``Netmamba: Efficient network traffic classification via pre-training unidirectional mamba,'' in \emph{2024 IEEE 32nd International Conference on Network Protocols (ICNP)}.\hskip 1em plus 0.5em minus 0.4em\relax IEEE, 2024, pp. 1--11.

\bibitem{peng2024ptu}
L.~Peng, X.~Xie, S.~Huang, Z.~Wang, and Y.~Cui, ``Ptu: Pre-trained model for network traffic understanding,'' in \emph{2024 IEEE 32nd International Conference on Network Protocols (ICNP)}.\hskip 1em plus 0.5em minus 0.4em\relax IEEE, 2024, pp. 1--12.

\bibitem{zhou2024trafficformer}
G.~Zhou, X.~Guo, Z.~Liu, T.~Li, Q.~Li, and K.~Xu, ``Trafficformer: an efficient pre-trained model for traffic data,'' in \emph{2025 IEEE Symposium on Security and Privacy (SP)}.\hskip 1em plus 0.5em minus 0.4em\relax IEEE Computer Society, 2024, pp. 102--102.

\bibitem{jacobs2022ai}
A.~S. Jacobs, R.~Beltiukov, W.~Willinger, R.~A. Ferreira, A.~Gupta, and L.~Z. Granville, ``Ai/ml for network security: The emperor has no clothes,'' in \emph{Proceedings of the 2022 ACM SIGSAC Conference on Computer and Communications Security}, 2022, pp. 1537--1551.

\bibitem{oh2023appsniffer}
S.~Oh, M.~Lee, H.~Lee, E.~Bertino, and H.~Kim, ``Appsniffer: Towards robust mobile app fingerprinting against vpn,'' in \emph{Proceedings of the ACM Web Conference 2023}, 2023, pp. 2318--2328.

\bibitem{guthula2023netfound}
S.~Guthula, R.~Beltiukov, N.~Battula, W.~Guo, and A.~Gupta, ``netfound: Foundation model for network security,'' \emph{arXiv preprint arXiv:2310.17025}, 2023.

\bibitem{wickramasinghe2025sok}
N.~Wickramasinghe, A.~Shaghaghi, G.~Tsudik, and S.~Jha, ``{ SoK: Decoding the Enigma of Encrypted Network Traffic Classifiers },'' in \emph{2025 IEEE Symposium on Security and Privacy (SP)}, May 2025, pp. 1825--1843.

\bibitem{geirhos2020shortcut}
R.~Geirhos, J.-H. Jacobsen, C.~Michaelis, R.~Zemel, W.~Brendel, M.~Bethge, and F.~A. Wichmann, ``Shortcut learning in deep neural networks,'' \emph{Nature Machine Intelligence}, vol.~2, no.~11, pp. 665--673, 2020.

\bibitem{han2024rules}
D.~Han, Z.~Wang, R.~Feng, M.~Jin, W.~Chen, K.~Wang, S.~Wang, J.~Yang, X.~Shi, X.~Yin \emph{et~al.}, ``Rules refine the riddle: Global explanation for deep learning-based anomaly detection in security applications,'' in \emph{Proceedings of the 2024 on ACM SIGSAC Conference on Computer and Communications Security}, 2024, pp. 4509--4523.

\bibitem{ye2024spurious}
W.~Ye, G.~Zheng, X.~Cao, Y.~Ma, and A.~Zhang, ``Spurious correlations in machine learning: A survey,'' \emph{arXiv preprint arXiv:2402.12715}, 2024.

\bibitem{steinmann2024navigating}
D.~Steinmann, F.~Divo, M.~Kraus, A.~W{\"u}st, L.~Struppek, F.~Friedrich, and K.~Kersting, ``Navigating shortcuts, spurious correlations, and confounders: From origins via detection to mitigation,'' \emph{arXiv preprint arXiv:2412.05152}, 2024.

\bibitem{nam2020learning}
J.~Nam, H.~Cha, S.~Ahn, J.~Lee, and J.~Shin, ``Learning from failure: De-biasing classifier from biased classifier,'' \emph{Advances in Neural Information Processing Systems}, vol.~33, pp. 20\,673--20\,684, 2020.

\bibitem{yang2024identifying}
Y.~Yang, E.~Gan, G.~K. Dziugaite, and B.~Mirzasoleiman, ``Identifying spurious biases early in training through the lens of simplicity bias,'' in \emph{International Conference on Artificial Intelligence and Statistics}.\hskip 1em plus 0.5em minus 0.4em\relax PMLR, 2024, pp. 2953--2961.

\bibitem{wang2023neural}
S.~Wang, R.~Veldhuis, C.~Brune, and N.~Strisciuglio, ``What do neural networks learn in image classification? a frequency shortcut perspective,'' in \emph{Proceedings of the IEEE/CVF International Conference on Computer Vision}, 2023, pp. 1433--1442.

\bibitem{nori2019interpretml}
H.~Nori, S.~Jenkins, P.~Koch, and R.~Caruana, ``Interpretml: A unified framework for machine learning interpretability,'' \emph{arXiv preprint arXiv:1909.09223}, 2019.

\bibitem{kokhlikyan2020captum}
N.~Kokhlikyan, V.~Miglani, M.~Martin, E.~Wang, B.~Alsallakh, J.~Reynolds, A.~Melnikov, N.~Kliushkina, C.~Araya, S.~Yan \emph{et~al.}, ``Captum: A unified and generic model interpretability library for pytorch,'' \emph{arXiv preprint arXiv:2009.07896}, 2020.

\bibitem{lapuschkin2019unmasking}
S.~Lapuschkin, S.~W{\"a}ldchen, A.~Binder, G.~Montavon, W.~Samek, and K.-R. M{\"u}ller, ``Unmasking clever hans predictors and assessing what machines really learn,'' \emph{Nature communications}, vol.~10, no.~1, p. 1096, 2019.

\bibitem{muller2024shortcut}
N.~M. M{\"u}ller, S.~Roschmann, S.~Khan, P.~Sperl, and K.~B{\"o}ttinger, ``Shortcut detection with variational autoencoders,'' in \emph{2024 International Joint Conference on Neural Networks (IJCNN)}.\hskip 1em plus 0.5em minus 0.4em\relax IEEE, 2024, pp. 1--7.

\bibitem{mink2023everybody}
J.~Mink, H.~Benkraouda, L.~Yang, A.~Ciptadi, A.~Ahmadzadeh, D.~Votipka, and G.~Wang, ``Everybody’s got ml, tell me what else you have: Practitioners’ perception of ml-based security tools and explanations,'' in \emph{2023 IEEE Symposium on Security and Privacy (SP)}.\hskip 1em plus 0.5em minus 0.4em\relax IEEE, 2023, pp. 2068--2085.

\bibitem{gil2016characterization}
G.~D. Gil, A.~H. Lashkari, M.~Mamun, and A.~A. Ghorbani, ``Characterization of encrypted and vpn traffic using time-related features,'' in \emph{Proceedings of the 2nd international conference on information systems security and privacy (ICISSP 2016)}.\hskip 1em plus 0.5em minus 0.4em\relax SciTePress, 2016, pp. 407--414.

\bibitem{HabibiLashkari2018TowardDA}
\BIBentryALTinterwordspacing
A.~H. Lashkari, A.~F.~A. Kadir, L.~Taheri, and A.~A. Ghorbani, ``Toward developing a systematic approach to generate benchmark android malware datasets and classification,'' \emph{2018 International Carnahan Conference on Security Technology (ICCST)}, pp. 1--7, 2018. [Online]. Available: \url{https://api.semanticscholar.org/CorpusID:56718203}
\BIBentrySTDinterwordspacing

\bibitem{wang2017malware}
W.~Wang, M.~Zhu, X.~Zeng, X.~Ye, and Y.~Sheng, ``Malware traffic classification using convolutional neural network for representation learning,'' in \emph{2017 International conference on information networking (ICOIN)}.\hskip 1em plus 0.5em minus 0.4em\relax IEEE, 2017, pp. 712--717.

\bibitem{ren2019international}
J.~Ren, D.~Dubois, and D.~Choffnes, ``An international view of privacy risks for mobile apps,'' 2019.

\bibitem{li2022robust}
W.~Li, X.-Y. Zhang, H.~Bao, Q.~Wang, and Z.~Li, ``Robust network traffic identification with graph matching,'' \emph{Computer Networks}, vol. 218, p. 109368, 2022.

\bibitem{dadkhah2022towards}
S.~Dadkhah, H.~Mahdikhani, P.~K. Danso, A.~Zohourian, K.~A. Truong, and A.~A. Ghorbani, ``Towards the development of a realistic multidimensional iot profiling dataset,'' in \emph{2022 19th Annual International Conference on Privacy, Security \& Trust (PST)}.\hskip 1em plus 0.5em minus 0.4em\relax IEEE, 2022, pp. 1--11.

\bibitem{lashkari2017characterization}
A.~H. Lashkari, G.~D. Gil, M.~S.~I. Mamun, and A.~A. Ghorbani, ``Characterization of tor traffic using time based features,'' in \emph{International Conference on Information Systems Security and Privacy}, vol.~2.\hskip 1em plus 0.5em minus 0.4em\relax SciTePress, 2017, pp. 253--262.

\end{thebibliography}

% \vspace{12pt}
% \color{red}
% IEEE conference templates contain guidance text for composing and formatting conference papers. Please ensure that all template text is removed from your conference paper prior to submission to the conference. Failure to remove the template text from your paper may result in your paper not being published.

\end{document}